\documentclass[runningheads,a4paper]{llncs}

\usepackage{amssymb}
\usepackage{graphicx}
\usepackage{amssymb}
\usepackage[utf8]{inputenc}
\usepackage{booktabs}
\usepackage{multirow}
\usepackage{array}
\usepackage{makecell}

\usepackage{url}
\urldef{\mailsa}\path|cwmokab@connect.ust.hk|
\urldef{\mailsb}\path|achung@cse.ust.hk|

\usepackage{amsmath}
\mathchardef\mhyphen="2D

\begin{document}

\mainmatter  % start of an individual contribution

\title{Learning Data Augmentation for Brain Tumor Segmentation with Coarse-to-Fine Generative Adversarial Networks}

\titlerunning{Learning DA for Brain Tumor Segmentation with GANs}

\authorrunning{Tony C.W. Mok and Albert C.S. Chung}

\author{Tony C.W. Mok and Albert C.S. Chung}
\institute{Lo Kwee-Seong Medical Image Analysis Laboratory,\\
Department of Computer Science and Engineering,\\
The Hong Kong University of Science and Technology, Hong Kong\\
\mailsa, \mailsb}

\maketitle

\begin{abstract}
There is a common belief that the successful training of deep neural networks requires many annotated training samples, which are often expensive and difficult to obtain especially in the biomedical imaging field. While it is often easy for researchers to use data augmentation to expand the size of training sets, constructing and generating generic augmented data that is able to teach the network the desired invariance and robustness properties using traditional data augmentation techniques is challenging in practice. In this paper, we propose a novel automatic data augmentation method that uses generative adversarial networks to learn augmentations that enable machine learning based method to learn the available annotated samples more efficiently. The architecture consists of a coarse-to-fine generator to capture the manifold of the training sets and generate generic augmented data. In our experiments, we show the efficacy of our approach on a Magnetic Resonance Imaging (MRI) image, achieving improvements of 3.5\% Dice coefficient on the BRATS15 Challenge dataset as compared to traditional augmentation approaches. Also, our proposed method successfully boosts a common segmentation network to reach the state-of-the-art performance on the BRATS15 Challenge.
%\keywords{Deep learning, Data augmentation, Generative Adversarial Networks}
\end{abstract}

\section{Introduction}
Accurate segmentation of a brain tumor from medical images is a crucial step for clinical diagnosis, evaluation, and follow-up treatment. Currently, the automatic segmentation methods which achieve state-of-the-art results are often using a deep learning approach. Modern deep learning models often consist of millions of parameters and learning these parameters requires massive annotated datasets to avoid overfitting to the training set.  However, the problem is made challenging by the number of annotated training datasets often being limited in the medical imaging domain due to a couple of reasons. First, it is time-consuming and expensive for experts to accurately delineate the pixel-wise brain tumor region. Second, manual labeling also suffers from considerable intra-rater and inter-rater inconsistencies \cite{menze2014multimodal}. Third, there are various modalities and imaging protocols, therefore a training set generated for one study is difficult to transfer to another study in practice.

To address these problems, we propose an automatic data augmentation approach for network-based brain tumor segmentation. Specifically, we present and evaluate a method for augmenting multimodal brain MRI images of high-grade (HG) and low-grade (LG) glioma patients, in which the generic augmented data enable the network-based method to learn the available annotated datasets more efficiently. Experimental results demonstrate that the proposed method effectively improves the segmentation accuracy of the network-based method, compared to the traditional data augmentation approach. It achieves improvements of 3.5\% dice coefficient on the BRATS15 Challenge dataset as compared to traditional augmentation approaches.
\section{Related Work}

\subsubsection{Data augmentation}
Data augmentation is essential to teach the network the desired invariance and robustness properties when only a few training samples are available. For medical image segmentation, different combinations of affine transformations are commonly used as data augmentation to teach the network the desired invariance and robustness properties. Ronneberger et al. \cite{ronneberger2015u} applied shift, rotation and elastic deformations to the microscopical images during training, while Milletari et al. \cite{milletari2016v} applied the random deformation to prostate MRI volumes using dense deformation field with B-spline interpolation. For brain tumor segmentation, scaling, rotation and flipping have also been applied to multimodal brain MR images for data augmentation \cite{shen2017boundary}. Typical data augmentation approaches fail to increase the diversity of the training data, i.e., different parameters for MR imaging protocol, tumor size, shape, location, and appearance. The contribution of this work is that we have developed an automatic way to learn a more generic augmentation so that not only the rotational and scaling invariance, high-level information such as the shape of tumor and contextual information can also be augmented.
\vspace{-9pt}
\subsubsection{Generative adversarial networks}
In the domain of computer vision, Generative adversarial networks (\textsc{GANs}) \cite{goodfellow2014generative} have elicited considerable attention. GANs aim to model the data distribution by forcing the generated sample to be indistinguishable from the data. They have also proven successful in a wide variety of applications such as image generation \cite{arjovsky2017wasserstein,rosca2017variational}, image manipulation \cite{zhu2016generative} and image inpainting  \cite{yeh2017semantic}. Recently, various coarse-to-fine frameworks of GANs have been proposed \cite{karras2017progressive,wang2017high} to generate high-quality and high-resolution images, e.g., $1024\times1024$ pixels. Inspired by their successes, we propose a new coarse-to-fine boundary-aware GANs suitable to generate generic augmented MR images for brain tumor segmentation.

\vspace{-10pt}
\section{Methods}
\begin{figure}[t]
	\centering
	\includegraphics[width=0.9\textwidth]{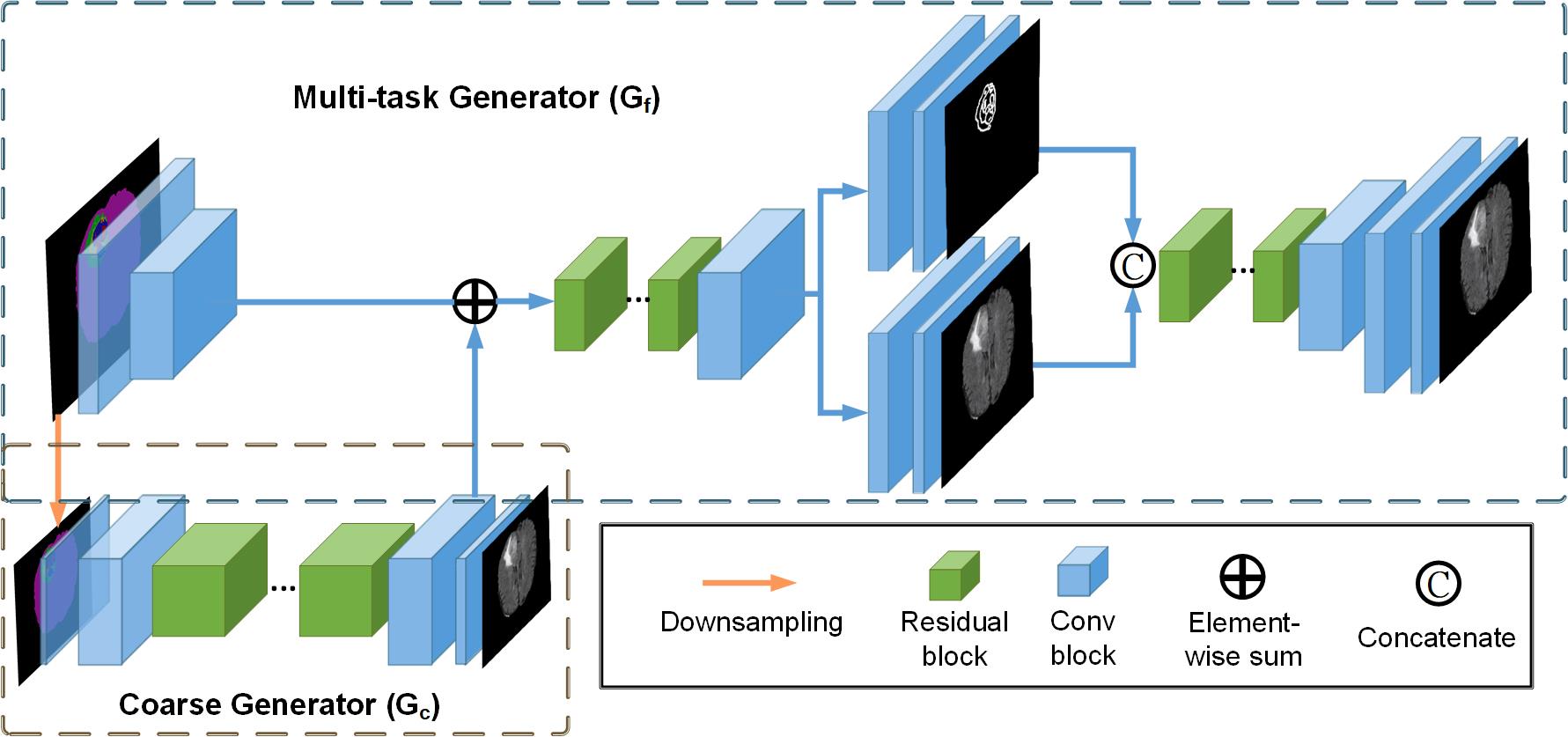}
	\caption{Network architecture of proposed generator.}
	\label{fig1}
	\vspace{-15pt}
\end{figure}
\subsection{Preliminaries of Generative Adversarial Networks}
Typical Generative Adversarial Networks (GANs), comprise a generator $G$ and a discriminator $D$ that are trained to compete with each other alternatively .  The generator $G$ is optimized to generate the data distribution $p_{data}$ by generating the images that are indistinguishable for the discriminator $D$ to differentiate from real images. While $D$ is optimized to distinguish real images and synthetic images generated by $G$. The training objective is similar to a two-player min-max game as follows:
\begin{equation}
\min_{G}\max_{D} \mathcal{L}_{GAN}(D,G) = \mathbb{E}_{x \sim p_{data}}[\log{D(x)}] + \mathbb{E}_{z \sim p_{z}}[\log{(1-D(G(z))}].
\end{equation}
\noindent where $x$ is a real sample from the target data distribution $p_data$, and $z$ is a noise vector sampled from distribution $p_z$.
\vspace{-10pt}
\subsection{Coarse-to-fine Boundary-aware Generator}
To generate high-resolution MR images of brains with realistic detail, we propose a Coarse-to-fine Boundary-aware Generative Adversarial Networks (CB-GANs). In our proposed method, the noise vector in traditional GANs is replaced by a label map of 2D axial slice from a 3D MR volume as a conditioning variable. We explain how we diversify the generated data without using a noise vector as input in Section 3.4.
\vspace{-10pt}
\subsubsection{Coarse-to-fine Generator}
Our generator is decomposed into two different sub-generators: $G_c$ and $G_f$. $G_c$ is the coarse generator while $G_f$ is the fine generator. The generator $G$ is then given by the tuple $G = \{G_c, G_f\}$. The coarse generator $G_c$ aims to sketch the primitive shape and texture of multimodal brain MR images from a label map at a lower resolution and the fine generator $G_f$ aims to correct the defects and completes the details of the low-resolution MR images from the coarse generator $G_c$.

The $G_c$ and $G_f$ consist of three components, namely, a convolutional downsampling block, a set of residual blocks, and a transposed convolutional block. The resolution of the input label map to $G_f$ is the same as the training data, while the resolution of the input label map to $G_c$ is $4 \times$ smaller than the training data (2x smaller along each axis). Different from the residual block in coarse generator $G_c$, the residual block in fine generator $G_f$ takes the element-wise sum of the output of $G_c$ and the input feature maps of previous layers from $G_f$ as the input. The element-wise sum operation helps integrate the global and local information from $G_c$ and $G_f$.
\vspace{-10pt}
\subsubsection{Boundary-aware Generator}
Although the above coarse-to-fine framework can already produce high resolution natural images, it remains a challenge to produce a high quality synthetic MR image of a brain tumor that serves the purpose of data augmentation, given the corresponding label map. Because the size of the tumor core in MR images is often small compared to the other encephalic regions, in which the networks may fail to notice that the details of the tumor core are important. Preserving accurate tumor boundaries is important for augmented data to teach the network the desired invariance and robustness properties. To address this problem, we propose a multi-task generator $G_f$ to replace the original fine generator.

The structure of the proposed generator is illustrated in Fig. 1. Instead of treating the image generation task as a single problem, we formulate it as a multi-task problem by exploring auxiliary information, which can simultaneously infer the location and boundary of the complete tumor. Specifically, two different branches are added to the final layer of $G_f$ in order to output the MR image of a brain with a tumor and the boundaries of the complete tumor. After that, the outputs from the two new branches are concatenated and fed into a residual block followed by a non-linear activation layer. Therefore, the boundary and texture information from the new branches are fused together to output the final image. The mean-square-error loss is used for the boundary extraction task, as shown in the following:
\begin{equation}
\mathcal{L}_b(x,y) = \frac{1}{n_i}\sum_n{\sum_i{(P(x_{n,i}; \theta) -  y_{n,i})^2}},
\end{equation}
\noindent where $\theta$ is the weight parameters in the generator. $\mathcal{L}_b$ refers to the mean-square-error loss for the boundary extraction task. $x_{n,i}$ and $y_{n,i}$ are the $i$-th pixel and ground truth in the $n$-th image used for training, respectively. $P$ refers to the predicted probability for the pixel $x_{n,i}$.
\vspace{-10pt}
\subsection{Adversarial training}
\subsubsection{Multi-discriminators}
As the resolution of the synthetic image increases, the difficulty for the discriminator to differentiate real and synthetic images also increase. When there is only a single discriminator, the discriminator needs to have a large receptive field that is able to capture both global, i.e., tumor location, and local, i.e., tumor texture and shape, information from the input image. However, this may not be a good idea as it implies we will need a discriminator which has either a deep network or large convolutional kernels. Both options require a large memory and may easily suffer from overfitting to the training data.

To address this challenge, we adopt multi-discriminators with different scales of input as our discriminator $D=\{ D_1, D_2, D_3, D_4 \}$. The four discriminators have identical architectures but operate at different image scales, which is similar to \cite{wang2017high}. Specifically, real MR images and synthesized MR images are downsampled by factors 2, 4 and 8 using the bilinear interpolation to create input for $D$ of 4 scales. Throughout the experiments, we find that using four discriminators can achieve the optimal performance and further increasing the number of discriminators cannot improve the quality of synthetic image.
\vspace{-10pt}
\subsubsection{Perceptual loss}
We further improve the GAN's loss by incorporating a modified perceptual loss. The main idea of the perceptual loss function is that if the synthetic image is similar to the real image, the $i$th-layer feature maps of the discriminator should be also similar when the synthetic and real images pass through it. The modified perceptual loss $\mathcal{L}_P(G, D_k)$ is calculated as:
\begin{equation}
\mathcal{L}_{P}(G, D_k) = \mathbb{E}_{(x, c)}\sum_{i=1}^{L}\frac{1}{N_i}[||D_k^{(i)}(x, c) - D_k^{(i)}(G(z), z)||_2^2],
\end{equation}
\noindent where $D_k^{(i)}$ represents the $i$th-layer feature maps of discriminator $D_k$, $L$ is the total number of layers, $N_i$ denotes the number of elements in each layer, $x$ denotes the real MRI image, $c$ denotes the label map and $z$ denotes the deformed label map.

Therefore, our full training objective combines both GANs loss and modified perceptual loss as:
\begin{equation}
\min_{G}\Big(\big(\max_{D_{k\in \{ 1,2,3,4\} } } \sum_{k=1,2,3,4} \mathcal{L}_{GAN}(D_k, G)\big) + \lambda_1 \mathcal{L}_b(x, y) + \lambda_2 \sum_{k=1,2,3,4} \mathcal{L}_{P}(G, D_k)\Big).
\end{equation}

\vspace{-12pt}
\subsection{Diversity of augmented data}
\begin{figure}[t]
	\centering
	\includegraphics[width=1.0\textwidth]{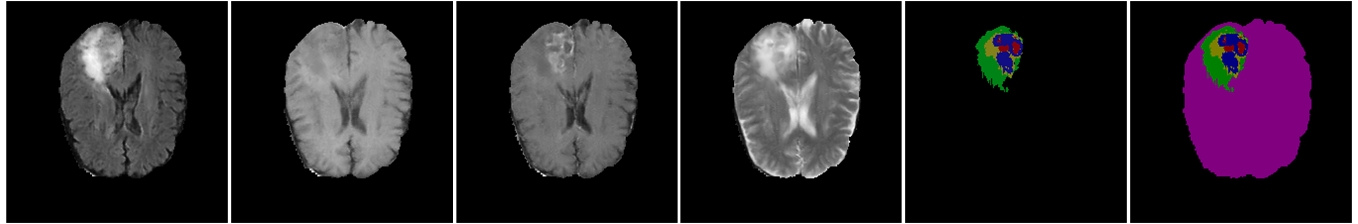}
	\caption{Example of a synthetic high-grade glioma tumor. Left to right: FLAIR, T1, T1c, T2, expert delineation (Ground truth) and semantic label map (Red: necrosis, Green: edema, Yellow: non-enhancing tumor, Blue: enhancing tumor).}
	\label{fig2}
	\vspace{-15pt}
\end{figure}

\subsubsection{Using deformed semantic label maps}
The traditional augmentation approach for object segmentation often uses different combinations of affine transformations, such as shifting, rotation, and zoom, to leverage the knowledge of invariances in a task. However, such knowledge implied by these affine transformations is limited.  For example, the shape, location and appearance of a complete tumor in a multi-modal MR image can vary significantly in the testing data, but the augmented image produced by the typical data augmentation fails to “simulate” such changes. Although some interpolation-based techniques such as elastic deformation can cause a slight variation in the shape of the augmented image, it may bring about damage and noise to the training data, as shown in fig. 3, if the deformation field varies a lot. 

Instead, we propose applying the elastic deformation to the label map. After that, we create a set of semantic labels from the deformed label maps. Specifically, we label $1$ to $5$ for necrosis, edema, non-enhancing tumor, enhancing tumor, non-tumor brain regions and $0$ for everything else in the semantic labels. We use the semantic labels instead of the label maps as input for our proposed CB-GANs. By providing the information of the contour of the brain to the generator, it further diversifies the synthetic brain MR image with different shapes and prevents model collapse, i.e., prevents the model from generating a set of realistic MR images with the identical shape and context of the brain. Figure 2 and 3 show the example of the synthetic image generated by CB-GANs with corresponding semantic label map.

\begin{figure}[t]
	\centering
	\includegraphics[height=4.8cm]{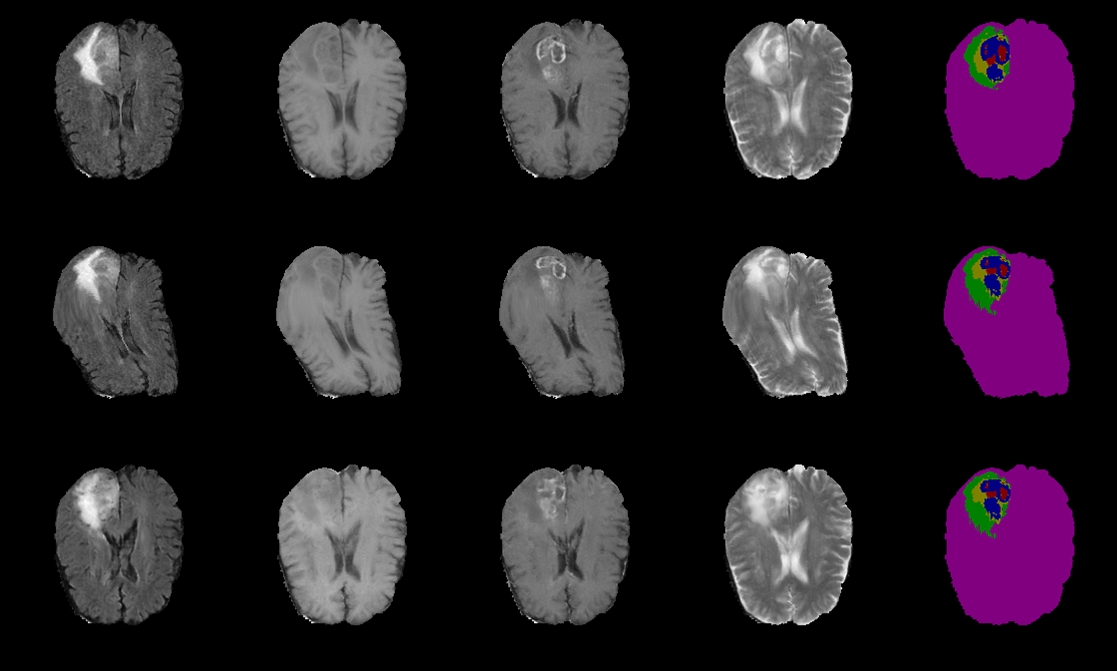}
	\caption{Comparison of traditional augmentation and our proposed method. First row: Original image. Second row: Augmented image using elastic deformation. Third row: synthetic image generated by our approach. Left to right: FLAIR, T1, T1c, T2 and semantic label map}
	\label{fig3}
	\vspace{-15pt}
\end{figure}

\vspace{-12pt}
\section{Experiments and Results}
\subsection{Data and Pre-processing}
Experiments have been performed using brain MRI sequences from BRATS15 datasets. The dimensions of each MR volume are $240\times240\times155$ pixels. BRATS15 provides both the training and test sets. The training set consists of 220 high-grade glioma (HGG) and 54 low-grade glioma (LGG) cases. For each case, it includes 4 modalities (Flair, T1, T1-contrast (T1c) and T2) that were skull-stripped and co-registered. Pixel-wise ground truths that annotate the complete tumor, which are verified by radiologists, are provided in the training set. For the testing set, it consists of 110 cases each with 4 modalities. Unlike the training set, the ground truth labels of the test cases are hidden from the public and evaluation is carried out via an online system. Extensive evaluation has been carried out on three tasks: (1) the complete tumor (necrosis, edema, non-enhancing and enhancing tumor) (2) the tumor core (necrosis, non-enhancing and enhancing tumor) (3) enhancing tumor region. For each MR image, we normalize the intensities of each modality to have zero-mean and unit variance. 

\vspace{-12pt}
\subsection{Network architectures}
\subsubsection{Generator architectures}
For generator networks, we adopt our backbone architectures from Wang et al. \cite{wang2017high} with some modifications. Below, we follow the naming convention used in the Wang et al. \cite{wang2017high}. Let $c7s1\mhyphen k$ denote a $7\times7$ Convolution-BatchNorm-ReLU layer with $k$ filters and stride 1. $dk$ denotes a $3\times3$ Convolution-BatchNorm-ReLU layer with $k$ filters and stride 2. $Rk$ denotes a residual block that contains two $3\times3$ convolutional layers with the same number of filters on both layers. $uk$ denotes a $3\times3$ fractional-strided-Convolution-BatchNorm-ReLU layer with $k$ filters and stride $\frac{1}{2}$. Note that we will replace the activation layer from ReLU to Tanh for the final layer of each generator.

\noindent Our coarse generator $G_c$:

\noindent$c7s1\mhyphen64$, $d128$, $d256$, $d512$, $d1024$, $R1024$, $R1024$, $R1024$, $R1024$, $u512$, $u256$, $u128$, $u64$, $c7s1\mhyphen4$

\noindent  Our fine generator $G_f$:

\noindent$c7s1\mhyphen32$, $d64$, $R64$, $R64$, $R64$, $u32$, $c7s1\mhyphen4$, $concat\{c7s1\mhyphen2, c7s1\mhyphen4\}$, $R64$, $R64$, $c7s1\mhyphen4$

\vspace{-12pt}  
\subsubsection{Discriminator architectures}
For discriminator networks, we use 4 Convolution-BatchNorm-LeakyReLU blocks for each discriminator. Let $Ck$ denote $4\times4$ Convolution-BatchNorm-LeakyReLU blocks with $k$ filters and stride 2. At the last layer, we add a sigmoid activation layer at the end to produce a 1-dimensional output. We use leaky ReLUs with default slope 0.2. All our four discriminators share the identical architecture as follows:

\noindent $C64$, $C128$, $C256$, $C512$

\vspace{-12pt}
\subsection{Network configuration and training}
During the experiments, we employ two sets of convolution neural networks (CNNs). The first set of CNNs is the proposed CB-GANs as shown in Fig. 1, which is used for generic data augmentation. While the second set of CNNs is the U-Net \cite{ronneberger2015u}, which is used for the segmentation task. We first trained the CB-GANs with back-propagation using the Adam optimizer with initial learning rate 0.0002 and momentum 0.5 for both generators and discriminators. We use CB-GANs to augment the training data during the training phase of U-Net. U-Net is trained with the same learning rate as CB-GANs. All the network are trained from scratch. The method is implemented using Pytorch. 

In terms of computation time, it takes about 4 days to train the CB-GANs and 20 hours to train the segmentation network for 100 epochs on a Nvidia GTX1080 Ti GPU. Moreover, we define the typical augmentation to be a combination of rotation (-10 to 10 degrees), zoom (0.98x to 1.02x) and random horizontal flip (50\%) that apply to the training data.
\vspace{-12pt}
\subsection{Evaluation}
We validate our approach by using it to augment the annotated training sets for the segmentation tasks and show that we have achieved strong gains, in terms of the Dice overlap metric between the automated segmentation and the radiologist annotation label map, over traditional augmentation baselines. We randomly split the training set in BRATS15 into two subsets, resulting in 234 training and 40 validation multimodal volumes. The full test set in BRATS15 is used as our test set, which includes 110 patients. 
\begin{table}[h]
	\centering
	\caption{Segmentation performance on the BRATS15 testing set. GANs: proposed architecture without coarse-to-fine framework and boundary loss function. C-GANs: coarse-to-fine GANs. CB-GANs: our proposed method.}
	\label{table1}
	\setlength\extrarowheight{3pt}
	\begin{tabular}{c|c|c|c|c|c|c|c|c|c}
		\hline
		\multirow{2}{*}{Method} & \multicolumn{3}{c|}{Dice} & \multicolumn{3}{c|}{Precision} & \multicolumn{3}{c}{Sensitivity} \\\cline{2-10}
		& Complete & Core & Enh. & Complete & Core & Enh. & Complete & Core & Enh. \\
		\Xhline{3\arrayrulewidth}
		GANs           & 0.80     & 0.58      & 0.55        & 0.84     & 0.80 & 0.62 & 0.80     & 0.55 & 0.51 \\
		\hline
		C-GANs         & 0.82     & 0.60      & 0.55        & \textbf{0.87}     & 0.80 & \textbf{0.66} & 0.81     & 0.55 & 0.52 \\
		\hline
		CB-GANs (ours) & \textbf{0.84}  & \textbf{0.63} & \textbf{0.57}  & \textbf{0.87} & \textbf{0.82}  & 0.65  & \textbf{0.84}  & \textbf{0.57}  & \textbf{0.54} \\
		\hline
	\end{tabular}
\end{table}
First, we conduct the component testing on the test set to evaluate the impact of coarse-to-fine framework and proposed boundary loss function. Table 1 compares the segmentation performance between a baseline GANs, a coarse-to-fine GANs and the proposed CB-GANs. It shows that if the coarse-to-fine framework and boundary loss function were added, there is an improvement in Dice values for the tumor core task, giving an average 3.6\% improvement in Dice. This is probably because the coarse-to-fine framework GANs and boundary loss function can correct defects and generate a clear boundary for small tumor regions in synthetic images. 

\begin{table}
	\centering
	\caption{Performance on the BRATS15 testing set. w/o DA: without any data augmentation. w/ DA: with typical data augmentation. w/ Proposed: with proposed generic data augmentation method.}
	\label{table2}
	\setlength\extrarowheight{3pt}
	\begin{tabular}{c|c|c|c|c|c|c|c|c|c}
		\hline
		\multirow{2}{*}{Method} & \multicolumn{3}{c|}{Dice} & \multicolumn{3}{c|}{Precision} & \multicolumn{3}{c}{Sensitivity} \\\cline{2-10}
		& Complete & Core & Enh. & Complete & Core & Enh. & Complete & Core & Enh. \\
		\Xhline{3\arrayrulewidth}
		w/o DA      & 0.79 & 0.54 & 0.43 & 0.85 & 0.79 & \textbf{0.66} & 0.78 & 0.47 & 0.37 \\
		\hline
		w/ DA       & 0.81 & 0.61 & 0.55 & 0.85 & \textbf{0.82} & 0.64 & 0.80 & 0.54 & \textbf{0.54} \\
		\hline
		w/ Proposed & \textbf{0.84}  & \textbf{0.63} & \textbf{0.57}  & \textbf{0.87} & \textbf{0.82}  & 0.65  & \textbf{0.84}  & \textbf{0.57}  & \textbf{0.54} \\
		\hline
	\end{tabular}
\vspace{-10pt}
\end{table}
We also compare the performance of the proposed method to the traditional augmentation method as listed in Table 2. Both data augmentation methods are able to improve the segmentation performance by a significant Dice value. Our proposed method further improves the performance over traditional data augmentation methods on average by 3.5\% of Dice values and achieves a significant improvement in Dice for the complete tumor task.

\begin{table}
	\centering
	\caption{Comparison to the state-of-the-art results on the BRATS15 testing set.}
	\label{table3}
	\setlength\extrarowheight{3pt}
	\resizebox{\textwidth}{!}{%
	\begin{tabular}{c|c|c|c|c|c|c|c|c|c}
		\hline
		\multirow{2}{*}{Method} & \multicolumn{3}{c|}{Dice} & \multicolumn{3}{c|}{Precision} & \multicolumn{3}{c}{Sensitivity} \\\cline{2-10}
		& Complete & Core & Enh. & Complete & Core & Enh. & Complete & Core & Enh. \\
		\Xhline{3\arrayrulewidth}
		Kamnitsasa17 \cite{KamnitsasLNSKMR17}         & \textbf{0.85} & 0.67 & \textbf{0.63} & 0.85 & \textbf{0.84} & 0.63 & \textbf{0.87} & 0.60 & 0.66 \\
		\hline
		Zhao17 \cite{ZhaoWSLZF18}               & 0.84  & \textbf{0.73} & 0.62  & \textbf{0.89} & 0.76  & 0.63  & 0.82  & \textbf{0.76}  & \textbf{0.67 } \\
		\hline
		2D U-net w/ proposed & 0.84  & 0.63 & 0.57  & 0.87 & 0.82  & \textbf{0.65}  & 0.84  & 0.57  & 0.54 \\
		\hline
	\end{tabular}
}
\end{table}
Finally, we compare our proposed method with two state-of-the-art methods as listed in Table 3. Kamnitsasa et al. \cite{KamnitsasLNSKMR17} method, achieving a top ranking in both BRATS15 and ISLES15 Challenge, using a dual pathway deep 3D CNNs to segment the tumor region and 3D fully connected Conditional Random Field to reduce the false positive, while Zhao \cite{ZhaoWSLZF18} joins three segmentation models which uses 2D image patches from different views. Our results are competitive with both methods and give better result for the enhancing tumor task in terms of Dice precision.

Also, one advantage of our proposed model is its relatively low computational cost in both the training and testing phases as we only leverage simple 2D CNNs with no post-processing method. Kamnitsasa et al. \cite{KamnitsasLNSKMR17} reported a running time of 3 minutes using a 3GB GPU to segment one case, while 6-12 minutes was reported by Zhao \cite{ZhaoWSLZF18}. With our proposed method, we achieve 2.1s for one case in inference time as the architecture of U-Net has much fewer learning parameters.

\vspace{-10pt}
\section{Conclusion}
In this paper, we propose a novel, automatic and network-based data augmentation method for brain tumor MR image segmentation. The main contribution is that we propose a generic way to augment training data that is able to teach network-based methods the desired invariance and robustness properties for segmentation tasks. We have shown that the proposed coarse-to-fine framework and boundary loss function in GANs lead to improved augmented data and segmentation quality. We have also shown that our method can boost a common segmentation network to reach the state-of-the-art multi-scale deep networks’ performance with the relatively low computational cost at inference time and outperforms the traditional augmentation methods. 

%\vspace{-12pt}
\bibliographystyle{splncs03}
\bibliography{miccai_ref}

\end{document}